\documentclass{article}

\usepackage{arxiv}

\usepackage[utf8]{inputenc} % allow utf-8 input
\usepackage[T1]{fontenc}    % use 8-bit T1 fonts
\usepackage{hyperref}       % hyperlinks
\usepackage{url}            % simple URL typesetting
\usepackage{booktabs}       % professional-quality tables
\usepackage{amsfonts}       % blackboard math symbols
\usepackage{nicefrac}       % compact symbols for 1/2, etc.
\usepackage{microtype}      % microtypography
\usepackage{lipsum}
\usepackage{graphicx}

\usepackage{IEEEtrantools}

\usepackage{subfigure}
\usepackage{latexsym}
\usepackage{amsmath}
\usepackage{xcolor}
\usepackage{xspace}
\usepackage{rotating}
\usepackage{dsfont}

\usepackage{akihito/deep_lasso}

\graphicspath{ {./images/} }

\title{LassoLayer: Nonlinear Feature Selection\\ by Switching One-to-one Links}
\author{
 Akihito Sudo \\
  Faculty of Informatics\\
  Shizuoka University, Japan \\
  \texttt{sudo@inf.shizuoka.ac.jp} \\
   \And
 Teng Teck Hou \\
  ST Engineering Ltd.\\
  Singapore\\
  \texttt{thteng@stengg.com} \\
  \And
 Masaki Yamaguchi \\
  \texttt{yamaguchi.masaki.17} \\
  \texttt{@shizuoka.ac.jp } \\
  \And
  Yoshinori Tone \\
  JAVIS CO., LTD.\\
   Vietnam. \\
  \texttt{tone@javis.vn} \\
}

\begin{document}

\twocolumn[{%
  \begin{@twocolumnfalse}
    \maketitle
    \begin{abstract}
      Along with the desire to address more complex problems, feature selection methods have gained in importance. Feature selection methods can be classified into wrapper method, filter method, and embedded method. 
Being a powerful embedded feature selection method, Lasso has attracted the attention of many researchers. However, as a linear approach, the applicability of Lasso has been limited.  
In this work, we propose LassoLayer that is one-to-one connected and trained by L1 optimization, which work to drop out unnecessary units for prediction.
For nonlinear feature selections, we build \LassoMLP: the network equipped with LassoLayer as its first layer.
Because we can insert LassoLayer in any network structure, it can harness the strength of neural network suitable for tasks where feature selection is needed. 
We evaluate  \LassoMLP in feature selection with regression and classification tasks.
\LassoMLP receives features including considerable numbers of noisy factors that is harmful for overfitting. 
In the experiments using MNIST dataset, we confirm that \LassoMLP outperforms the-state-of-the-art method. The speriority of LassoMLP is significant especially when the number of the training data is small.
% In experiment 1111, we show {\color{blue}abc abc abc abc abc abc abc abc abc abc abc abc abc abc abc abc abc abc abc abc abc abc abc abc abc abc abc abc abc abc abc abc abc abc abc abc abc}.
% In Experiment 2, the task is classifying 1 or 7 from hand-writing digits.
% Surprisingly, on the classification with very sparse dataset (training 17 samples from MNIST786 adding 5000 noisy features, i.e. \#dim is 5786), \LassoMLP achieves 95\% accuracy. Furthermore, 100\% noise features are dropped and only 4 of 786 pixels are selected to achieve this accuracy. Overfitting suffers baselines, and thus, those accuracies are less than XX\%.

      \\[10pt]
    \end{abstract}
    
  \end{@twocolumnfalse}
}]

% keywords can be removed

\section{Introduction}
\label{sec:introduction} Due to advent of sensor technologies, data can be gathered at ever higher resolution and better precision. Thus, there is the desire to gain invaluable insights from high-dimensional data. A technique called sparse modeling is gaining attention as a method for extracting information from high-dimensional data. Using the sparsity of high dimensional data, sparse modeling can perform calculations in realistic amount of time. Even in situations where the amount of calculation exponentially increases with respect to the number of dimensions, it narrows down to make it easy to extract rules from data.

Lasso \cite{lasso} is often used to implement sparse modeling. Using Lasso, it is possible to obtain a model expressing the input / output relation after choosing only the required dimensions from the high dimensional data. It can also be regarded as a method of selecting characteristics of data consisting of a set of explanatory variables \cite{embedded} classified as an embedded feature selection method.

Lasso is a linear regression model in which the loss function has an L1 regularization term and has the property that it is easy to obtain a sparse solution such that the coefficient of the linear regression model becomes zero with the effect of the L1 regularization term. Due to this effect, it becomes possible to construct a model with as few explanatory variables as possible for data of high dimensional inherent in sparseness that many dimensions of explanatory variables do not affect the explained variable. Various Lasso extensions have been studied so far, Group Lasso \cite{group} which collectively tends to have a coefficient of zero in group units, Fused Lasso \cite{fused} which can consider spatio-temporal adjacency.% Is famous.

% \textcolor{red}{I think this paragraph is too long. Maybe, most part of this paragraph should be moved to section 2 (related works).}

\begin{figure}[tb]
\centering
\subfigure[Training Step]{
  \centering
       \includegraphics[clip, width=.44\columnwidth]{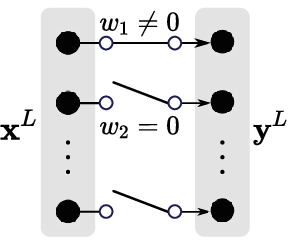}%
}
\quad
\subfigure[Prediction Step]{
  \centering
       \includegraphics[clip, width=.44\columnwidth]{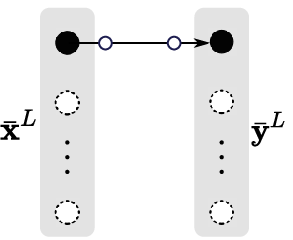}%
}
       \caption{The architecture of LassoLayer.}
         \label{fig:method}
\end{figure}

Lasso-based approaches such as standard Lasso, Group Lasso and Fused Lasso are known to not represent non-linear input-output relationships. Roth et al. \cite{instance-wise} suggested to convert variable by nonlinear function to explanatory variable and obtain linear regression model with Lasso for vector obtained by transformation. However, in this method, since features are selected for the converted vector, it is not possible to select the feature of the explanatory variable of the original data. Li et al. \cite{feature-wise} proposed a method called Feature Vector Machine (FVM) as a nonlinear method that can perform feature selection on the original explanatory variable. In the method of Roth et al., non-linear functions are used to convert explanatory variables, whereas in FVM, non-linear functions are used to convert vectors comprising explanatory variables. As a result, features are selected for each dimension of the original explanatory variable. Later, Yamada \cite{yamada} et al. found that, irregardless of the types of problem, FVM was not able to obtain a good solution when the number of samples falls below the dimension number or when explanatory variable is converted using a nonlinear function. Thus, we propose a method to convert explanatory variables and explanatory variables by using the kernel suggested by \cite {stein}, because of the limitation of flexibility to capture non-linear relationships among the explanatory variables.
% The kernel function is not necessarily positive definite kernel

% On the other hand, research to nonlinearize Lasso using kernel method has been reported \ cite {xxx, xxx}. Yamada et al. Reported nonlinearity by XXXX and it was also effective for Non additive model.

Meanwhile, architectures with deep networks achieve SOTA in various fields \cite{deep}. 
In order to improve the efficiency and accuracy of deep learning,
researchers has been proposed
various techniques such as convolution layer and pooling layer network structure such as \cite{convnet}, gradient method \cite{adam, adadelta}, batch learning method \cite{batchnorm} .
Because feedforward neural networks of three or more layers can approximate arbitrary nonlinear functions, if this deep learning can be applied to Lasso, it is considered that an embedded type feature extraction method combining nonlinearity and deep learning performance can be obtained.

\begin{figure}[tb] % picture
    \centering
    \includegraphics[width=.8\linewidth]{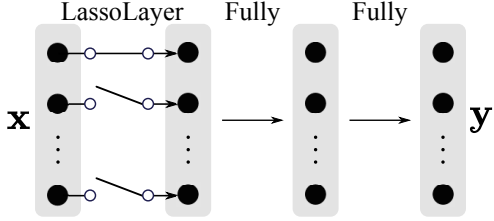}
    \caption{LassoMLP}
    \label{fig:LassoMLP}
\end{figure}

In the present paper, we present \LassoLayer, which is designed as a filter to achieve feature selection.
\LassoLayer performs feature selection through passing only the features that are effective for prediction to the subsequent layer.
Figure~\ref{fig:method} depicts the architecture of \LassoLayer, where the same number of units are connected by one-to-one links.
Since the weights are optimized by L1 regularization, the links act as switches that pass only the effective features to the subsequent layers.
By placing \LassoLayer at the head of the whole network, feature selection for raw inputs is achieved whilst feature selection for the representations in hidden layers can be realized by placing \LassoLayer in between the hidden layers.
Figure~\ref{fig:LassoMLP} shows one of the simplest implementations for feature selection, which we will elaborate on in Section 3.

The presentation of this work continues in Section~\ref{sec:related_work} where a collection of related works are described and contrasted. This is followed by the presentation of our proposed Deep-Lasso method in Section~\ref{sec:LassoLayer}. Experiments conducted to evaluate and compare the performance of our proposed Deep-Lasso method are described in Section~\ref{sec:experiment}. The proposed approach is evaluated using key performance indicators such as accuracy of feature selection and approximation accuracy of the function. Last but not least, Section~\ref{sec:conclusion} concludes and summarises this work.

\section{Related Work}
\label{sec:related_work} This section will compare and contrast this work with seminal works addressing the problem of feature selection~\cite{girish2014}. In particular, the survey is steered towards the use of deep neural network for feature selection and as a Least Absolute Shrinkage and Selection Operator (Lasso).

% Part 1: Survey 2 - 3 seminal works on feature selection
Feature selection, also known as variable elimination, is a necessary pre-requisite to many machine learning problems~\cite{girish2014}. There are the Filter, Wrapper and Embedded methods. Variable ranking techniques are used as the principle criteria for variable selection in the \emph{filter} methods. A subset of variables is determined using performance of predictor in the \emph{wrapper} methods. The process of eliminating variables is incorporated as part of the training process in the \emph{embedded} methods. Common classifiers used for feature selection tasks are support vector machines (SVM)~\cite{vapnik1992} and radial basis function (RBF)~\cite{broomhead1988}. Variants of the cross validation method method are used for validating the performance of the classifiers calibrated using the selected features.

From a different perspective, feature selection algorithms (FSAs) can be represented using a set of characteristics comprising search organization, generation of successor states and evaluation measures~\cite{molina2002feature}. Feature selection methods such as the Las Vegas Filter (LVF), Las Vegas Incremental (LVI) algorithm, the Relief algorithm, the Sequential Forward Generation (SFG) algorithm, Sequential Backward Generation (SBG) algorithms, Sequential Floating Forward Search algorithm, the Focus algorithm, the Branch $\&$ Bound (BB) algorithm and the Quick BB (QBB) algorithm are represented using the suggested approach. It is also suggested in \cite{molina2002feature} that FSAs be evaluated with respect to particularities such as the relevance, irrelevance, redundance and sample size of the selected features. Four instances from each of the three classical problems are used to generate datasets for evaluating the FSAs. The three classical problems are the \emph{parity} problem, the \emph{disjunction} problem and the \emph{GMonks} problem.

Recently, researchers proposed feature selection methods for nonlinear dataset. HSIC Lasso is a nonlinear Lasso using kernel method \cite{yamada}. LassoNet selects features by back propagation of a network with residual connection in a network \cite{LassoNet2021}. We will compare our method with these methods in the experimental section.

% A feature selection challenge conducted during NIPS 2003 conference offered a baseline reference point to the performance of FSAs~\cite{guyon2005result}. The ARCENE, DEXTER, DOROTHEA, GISETTE and MADELON datasets are used during this feature selection challenge. The submitted entries are assessed using metrics comprising the Balanced Error Rate (BER), the Area under the ROC curve (AUC), Fraction of features selected (Ffeat) and Fraction of probes found in feature set selected (Fprobe). The methods as seen from the $56$ eligible submissions belong to coarse categories such as feature selection, classifier, ensemble methods, transduction and pre-processing. It is concluded from this feature selection challenge that \emph{"feature selection can be performed effectively"} and \emph{"eliminating meaningless features is not critical to achieve good classification performance"}.

% Part 2: Survey 2 - 3 seminal works on the use of Lasso for feature selection

% Part 3: Survey 2 -3 seminal works on the use of deep neural network for feature selection and as non-linear Lasso.

\section{LassoLayer and LassoMLP}
\label{sec:LassoLayer}
In this section, after illustrating how \LassoLayer works, we introduce \LassoMLP, which equips \LassoLayer at the first layer. We utilize \LassoMLP to evaluation the performance of \LassoLayer in the following section.

\subsection{Architecture of LassoLayer}
The $\LassoLayer$ consists of two layered units, those are linked one-by-one as illustrated in Fig.~\ref{fig:method}.
A bias unit is not necessary.
Let $\nLL$ be the number of the $\LassoLayer$'s input units.
There are the same number of the output-side unites.
We denote by $\wLL$ be the weights in the LassoLayer.
The weight $\wLL$ is a $\nLL$ dimensional vector.

Suppose that the \LassoLayer receives the inputs $\xLL$.
The input-side unites activates the inputs by $\sigmaLLIN$, and pass them to the output-side units through the weighted one-to-one link; The output units receive the vector $\wLL \odot \sigmaLLIN(\xLL)$, where we denote Hadamard product by $\odot$.
Finally, the output units activate them, and thus, the $\LassoLayer$ produces the $\nLL$ dimensional output $\yLL$:
\begin{IEEEeqnarray}{lCr}
\yLL = \sigmaLLOUT\left(\wLL \odot \sigmaLLIN(\xLL)\right).
\IEEEeqnarraynumspace
\label{eq:lassoLayerInOut}
\end{IEEEeqnarray}

\subsection{Training of LassoLayer}
\label{training}
For simplicity, we assume that there is exactly one \LassoLayer in the network to illustrate how the network is trained although it is straightforward to extend the training scheme to the network equipped with more than two \LassoLayer.
Let $\theta$ be the trained parameter except for $\wLL$.
We denote by $\otherLossFunc$ the loss function for the network when the network does not contain \LassoLayer.

To realize the feature switching, we add a $L_1$ penalty of $\wLL$ to a loss function:
\begin{IEEEeqnarray}{lCr}
\lossFunc = \otherLossFunc + \lambda \|\wLL\|_1
\IEEEeqnarraynumspace
\end{IEEEeqnarray}
where $\lambda$ is a hyperparameter that determines the size of the L1 penalty.
Since the derivative of the penalty term vanishes, only the first term affects the gradient to update the parameters $\theta$.
Meanwhile, the weights $\wLL$ in \LassoLayer is trained with the L1 penalty, which shrinks less important inputs to zero.
The zero weights filters the corresponding weights. 
Thus, the one-to-one links in \LassoLayer are suppose to switch the inputs by the penalty term.
Traditional regularization techniques often employ the $L_1$ norm or $L_2$ norm of all parameters as a regularization term, but $L_1$ norm of only $\LassoLayer$'s weights is required to achieve feature selection by $\LassoLayer$

For $\otherLossFunc$, one can select any loss function known effectively working for the whole network such as mean square error, cross entropy. Additional regularization term is also permissible.

We introduce a heuristic for stable training; kicking out the zero-valued weights to become non-zero during in the early phase of training.
To prevent false negative on feature selection, the zero-valued weights in \LassoLayer are kicked out to inspect whether the corresponding features are really ineffective for a prediction.
Precisely, during the first $\kickEpoch$ epoch, the zero-valued weights in $\LassoLayer$ are updated to have some non-zero values by the probability of $\kickProbability$. The updated values are chosen from either $-\kickedValue$ or $\kickedValue$ by the probability 0.5.
Since these strategies are found by empirical observations and increase the effort of hyperparameter tuning, finding the theoretically optimal strategies is a important direction for future work.

% We introduce two heuristics for stable training; initializing the weights $\wLL$ with very small values; kicking out the zero-valued weights to become non-zero during in the early phase of training.
% Firstly, we empirically discovered that tiny initial values of $\wLL$ drastically stabilize feature selection performance. Perhaps, this is because less complex model can avoid overfitting.
% We sample the initial weights of \LassoLayer from the uniform distribution in $[-\initWeightWidth, \initWeightWidth]$.
% We empirically find the training is stable when $\initWeightWidth=0.01$.
% Secondly, to prevent false negative on feature selection, the zero-valued weights in \LassoLayer are kicked out to inspect whether the corresponding features are really ineffective for a prediction.
% Precisely, during the first $\kickEpoch$ epoch, the zero-valued weights in $\LassoLayer$ are updated to have some non-zero values by the probability of $\kickProbability$. The updated values are chosen from either $-\kickedValue$ or $\kickedValue$ by the probability 0.5.
% Since these strategies are found by empirical observations and increase the effort of hyperparameter tuning, finding the theoretically optimal strategies is a important direction for future work.

Further, it is known that some heuristics is required for making parameters exactly zero by $L1$ penalties.
Without such heuristics, even if loss functions have the $L_1$ term , the trained parameter is unlikely to become zero. 
Hence, in practice, we employ a method such as Duchi's method \cite{fobos} to realize zero-valued parameters. 
In Duchi's method, when the absolute value of a parameter is smaller than the hyperparameter $\lambda$, the parameter value is forced to become zero. %$$, and the aim of the proposed method that the coupling weight corresponding to the unnecessary dimension of the feature amount becomes zero is easily realized.

% Furthermore, despite being originally effective for estimation, the value of the coupling weight may become zero for reasons such as the small absolute value of the initial value and the like. To prevent this, we add either $ \Delta w $ or $ - \Delta w $ to the coupling weight of the first layer with zero probability $ \rho $ at the initial stage of learning. This is done with the first $ m_1 $ epoch to do the learning and not with the next $ m_2 $ epoch. $ \rho $, $ \Delta w $, $ m_1 $, $ m_2 $ are hyperparameters and $ m_1 + m_2 $ is equal to the total number of epochs for learning.

\subsection{Prediction with LassoLayer}

There are multiple ways to make predictions with a network equipped with LassoLayer. In this section, we introduce two methods. One is to make predictions on the raw network that has been trained, and the other is to make predictions using only the features corresponding to the top-k weights of the LassoLayer.
For the former method, the predictions of the network are given through computing the output of LassoLayer with the equation (\ref{eq:lassoLayerInOut}).
Let $w^k$ and $w_i$ be the $k$-th largest element of $\mathbf{w}$ and $i$-th element of $\mathbf{w}$, respectively.
For the latter method, the output of LassoLayer is determined as follows:
\begin{IEEEeqnarray}{lCr}
\yLL = \sigmaLLOUT\left(\vec{\mathds{1}}_{\geq w^k} \odot \sigmaLLIN(\xLL)\right),
\IEEEeqnarraynumspace
\label{eq:lassoLayerPredictOut}
\end{IEEEeqnarray}
where $\vec{\mathds{1}}_{\geq w^k}$ is the $n$-dimensional vector of which $j$-th element is determined by
\begin{IEEEeqnarray}{lCr}
\mathds{1}^i_{\geq w^k}= 
\begin{cases}
1 & \text{if\ } w_i \geq w^k\\
0 & \text{otherwise}.
\end{cases}
\IEEEeqnarraynumspace
\end{IEEEeqnarray}

\subsection{LassoMLP}
\label{sec:LassoMLP}
To evaluation \LassoLayer, we implement the network as the sequence of a \LassoLayer and a three-layered MLP, as illustrated in Fig.~\ref{fig:LassoMLP}.
We call it \LassoMLP and believe this architecture is the simplest one for \LassoLayer to achieve the nonlinear feature selection of the input data.
The \LassoLayer receives and filters raw inputs to pass only effective features to MLP.
The MLP performs a nonlinear transformation on the filtered data.
passed to the \LassoLayer, and filtered to pass through 
the filtered data are processed.

Suppose that the \LassoMLP receives the inputs $\netIn\in\mathbb{R}^{\Nin}$ and produces the outputs $\netOut\in\mathbb{R}^{\Nout}$.
Both the activation functions $\sigmaLLIN$ and $\sigmaLLOUT$ are the identity map.
Thus, by Eq.~(\ref{eq:lassoLayerInOut}), the \LassoLayer passes the filtered signal $\yLL=\wLL \odot \netIn$.
We denote by $\Nhidden$ the number of the hidden units of the MLP. 
Let the weights of first and second layer in the MLP be $\WmlpO$ and $\WmlpT$.
The weights $\WmlpO$ and $\WmlpT$ are $\Nin\times\Nhidden$ and $\Nhidden\times\Nout$ matrices.
The outputs $\netOut$ are determined as follows:
\begin{IEEEeqnarray}{lCr}
\netOut = \sigmaOut\left[\WmlpT\,\sigmaHidden\left(\WmlpO \yLL\right)\right]
\IEEEeqnarraynumspace
\end{IEEEeqnarray}

\begin{figure}[tb] % picture
    \centering
    \includegraphics[width=.9\linewidth]{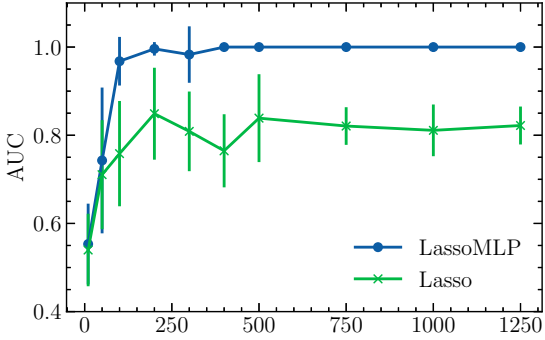}
    \caption{Feature selection performance with the synthetic dataset.}
    \label{fig:Synthe_Nonadditive_AUC}
\end{figure}

\begin{figure*}[tb] % picture
    \centering
    \includegraphics[width=.95\linewidth]{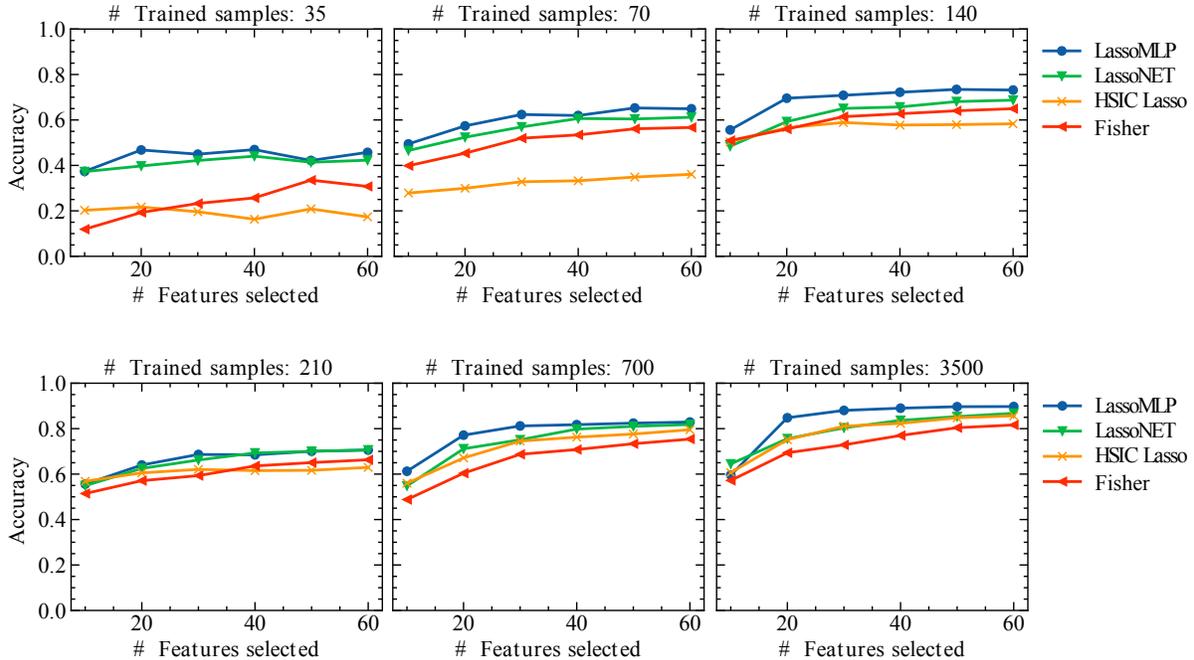}
    \caption{Feature selection performance with MNIST dataset.}
    \label{fig:FS_performance}
\end{figure*}

% \begin{figure}[tb] % picture
%     \centering
%     \includegraphics[width=.9\linewidth]{img/deep_lasso/mnist_13.eps}
%     \caption{MNIST hand written digit images.}
%     \label{fig:mnist13_images}
% \end{figure}

\section{Experiments}
\subsection{Experiments with Nonlinear Synthetic Data}
\label{sec:experiment}

\subsubsection{Setting}
\label{setting}

% \begin{figure*}[tb]
% \centering
% \subfigure[Additive model]{
%       \includegraphics[clip, width=.8\columnwidth]{img/deep_lasso/ibisml_result/time_additive.eps}}%
% \subfigure[Non-additive Model]{
%       \includegraphics[clip, width=.8\columnwidth]{img/deep_lasso/ibisml_result/time_non_additive.eps}}%
%       \caption{The process time for the training step}
%          \label{fig:time}
% \end{figure*}

To confirm that the proposed method can perform feature selection and function approximation on non linear data, we compare LassoMLP with Lasso using synthetic dataset.
The number of dimensions $ P $ of input data is 256 dimensions, and the dimension of output is 1. 
That is, the data $D$ is a set of  the pairs $(\mathbf{x}, y)$, where $ \mathbf{x} \in \mathbb {R}^{256} $ and $ y \in \mathbb{R}$. Each element of $ \mathbf{x} $ is generated from the standard normal distribution of which the mean is 0 and the variance is 1.
To generate the value of $y$ corresponding to $\mathbf{x}$, we utilized the following nonlinear functions:
\begin{eqnarray}
y = \sin(x_1) \exp(-x_2) + (x_3-0.2)^2,
\label{eq:non_ad}
\end{eqnarray}
where $(x_1,x_2,x_3)$ is the first three elements of $\mathbf{x}$.
We denote the number of the pairs in $D$ by $N$.
%In this experiments, $N$ is set to one of $ [76, 128, 179, 256, 384, 512] $.

%The number of layers in the proposed method was four, with the first multilayer perceptron consisting of single bonds and the second and subsequent layers consisting of three layers.
The number of of hidden units was 100.
The hyperparameter of LassoMLP is as follows: $\lambda=0.0001$, $ \rho=0.1 $, and $ \delta = 0.1 $.
For Lasso, $\lambda=0.1$.
The optimizer was the stochastic gradient descent (SGD) with the learning rate 0.1.
In addition, we use the mini batch learning with the batch size determined by $\min(80, D)$.
The total number of training epoch is 6000. Of those epochs, the kicking strategy was applied in the first 1000 epochs; namely $\kickEpoch=1000$.
%Learning times were 1,400 epochs, with $ m_1 $ as 1000 and $ m_2 $ as 400.
The activation function of the hidden units is ReLU.

To evaluate the feature-selection performance, we measured AUC score.
% We regard that the relevant features (i.e. $x_1$ to $x_3$ ) are positive samples and other features are negative samples, and the models solve this binary classification task.

% For the evaluation index, F1 value was used for feature quantity selection, and the root mean square error (RMSE) was used for function approximation performance.
% The F1 value was calculated assuming that the dimension effective for output is a positive example and the other dimension is a negative example. For evaluation of function approximation performance, cross validation of 10-fold was performed and RMSE was calculated for test data.

% The F1 value was calculated with the setting of the classification task where the effective features are  positive  and non-effective features are negative. For evaluation of function approximation performance, cross validation of 10-fold was performed.

% The baseline method is Lasso, of which the loss function is
% \begin{eqnarray}
% \frac{1}{2N}||\textbf{y} - X\textbf{w}||^2_2 + \lambda ^{Lasso} ||\textbf{w}||_1
% \end{eqnarray}
% where $ N $ is the number of data, $ Y = y^T$ and $ X = x^T (X \in \mathbb{R}^{N \times 256} $. $\textbf{w}$ is the coefficient of the linear regression model. $\lambda^{Lasso}$ is the hyper parameter affecting the strength of the $L_1$ penalty term. In this experiment we set $\lambda ^{Lasso} = 1.0$.

\subsubsection{Result}
Fig.~\ref{fig:Synthe_Nonadditive_AUC} shows ACU of the feature selection task for varying number of the training data.
While Lasso and LassoMLP are competitive when the number of the training data is small, LassoMLP outperforms Lasso when the training data is greater than 100. Then, although AUC of Lasso is approximately 0.8, that of LassoMLP is approximately 1.0.
%For the reader's convinience we draw the black dotted line in Fig.~\ref{fig:f1} that indicates the number of the data equal to the number of dimensions of the input (i.e. $ N = P $).
% For both the feature selection and function, LassoMLP clearly outperforms Lasso.
% The F1 score of LassoMLP is over 0.65 on $ N \geq P $ (that is, 256) while that of Lasso is approximately $ 0.1 $ regardless of the number of data. Regarding the function approximation task,  while the RMSE of LassoMLP was approximately $ 2.0 $, that of Lasso was more than $4.0$.
This result suggests that LassoMLP performs better than the vanilla Lasso for the nonlinear dataset if sufficient dataset is provided.
%For non-additive models, LassoMLP was 0.8 to 2.2, Lasso was between $ 4.1 $ and $ 5.6 $.

% Figure \ref{fig:time} shows the sum of the processing time of learning and estimation of the proposed method. The number of data was 512, which was 43.4 seconds and 45.6 seconds in the additive model and the non-additive model respectively, and processing time roughly linearly increased with the number of data within the scope of this experiment.

\subsection{Feature Selection Performance with Real-world Dataset}
\label{sec:Experiments 10 with MNIST}

We compare the performance of LassoMLP with baseline methods in terms of feature selection.

\subsubsection{Setting}

The task is the feature selection from MNIST data so as to classify the digits accurately with fewer features. The dimension of the images is 784 (28 x 28). We evaluated LassoMLP and baseline methods with the classification accuracy with the selected features. 
Firstly, the features are selected by each method using only training data. Secondly,  Random Forest (RF) are trained by the training data including only selected features. Lastly, the accuracy is evaluated by the test data that also include only selected features.
We adopt this evaluation scheme by following Ref.~\cite{LassoNet2021}.

We added the additional non-relevant features to MNIST images. The number of dimension is 5000. The non-relevant features are sampled from the standard Gaussian. The dataset was divided randomly into the training data and the test data with varying split size from 0.05\% to 5\% training samples. Then, while the number of dimension is 5784, the number of the training samples is from 35 to 3500. The variation of the numbers of the trained samples are shown in the title of Fig.~\ref{fig:FS_performance} (i.e. $35, 70, 140, \cdots, 3500$).
We employed this unbalance split because we are interested in the feature selection in sparse dataset.
The validation data is not necessary because we used the default set of the hyper parameters of SVM and RF was employed across all the baselines.

The hyper parameters of LassoMLP are as follows: batch size is 80, $\lambda = 0.005$, $\rho = 0.1$, $\delta = 0.25$. The optimizer is SGD with the learning rate 0.1.

The baseline methods include LassoNet, HSIC Lasso, PFA, and Fisher score. We follows Ref.~\cite{LassoNet2021} to choose the baselines.
For LassoNet and HSIC Lasso, we made use of the implementation in their original github repositories.
We used the default hyper paramters of LassoNet and HSIC Lasso in their implementations.
We implemented PFA using the implementations of PCA and k-means in scikit-learn. We used the implementation of Fisher score in scikit-feature.

\subsubsection{Results}
Fig.~\ref{fig:FS_performance} shows the classification accuracy by RF using only selected features with varying training samples. The selected features varies from 10 to 60.
%When 60, 30, 20, 10 features were selected, the accuracies of SVC for (LassoMLP (ours), LassoNet, and HSIC Lasso) are (\textbf{84.9}\%, 84.7\%, 81.7\%), (\textbf{83.6}\%, 77.1\%, 75.9\%), (\textbf{79.4}\%, 72.6\%, 67.2\%) and (\textbf{60.4\%}, 55.8\%, 55.8\%). Likewise, with RF, (\textbf{82.4}\%, 81.6\%, 79.2\%), (\textbf{81.0}\%, 75.2\%, 74.3\%), (\textbf{78.2}\%, 71.5\%, 67.1\%) and (\textbf{60.9}\%, 56.7\%, 55.7\%).

LassoMLP outperformed all baseline methods for all the number of features and trained samples at least in this experimental setting with two exceptions, the pairs of (\# trained sample and \# features) are (210, 40) and (3500, 10).
% The differences of accuracies between LassoMLP and other baselines, LassoNet and HSIC Lasso, are significant when less features are selected.

\begin{table*}[t]
\centering
\caption{Accuracy on MNIST784+Gauss5000 for the average (variance) of running experiments 10 times.}
\label{tbl:accuracy_MNIST_2classes}
\subfigure[Classify 1 from 4\label{tbl:accuracy_MNIST1-4_table}]{
    \begin{tabular}{rllll}
    \toprule
    the number of training data &                  4 (0.03\%)  &                  7 (0.05\%)  &                  15 (0.1\%) &                  75 (0.5\%)\\
    \midrule
    Multilayer Perceptron      &  0.512 ($\pm$0.021) &    0.481 ($\pm$0.0) &  0.558 ($\pm$0.166) &  0.729 ($\pm$0.222) \\
    Random Forest       &   0.501 ($\pm$0.02) &  0.526 ($\pm$0.041) &  0.704 ($\pm$0.152) &  \textbf{0.975} ($\pm$0.009) \\
    Support Vector Classifier      &  0.627 ($\pm$0.146) &  0.573 ($\pm$0.119) &    0.8 ($\pm$0.178) &  0.967 ($\pm$0.009) \\
    \textbf{LassoMLP (ours)} &  \textbf{0.744} ($\pm$0.083) &  \textbf{0.692} ($\pm$0.079) &  \textbf{0.941} ($\pm$0.006) &  0.955 ($\pm$0.002) \\
    \bottomrule
    \end{tabular}
}%
\quad
\subfigure[Classify 1 from 7  
\label{tbl:accuracy_MNIST1-7_table}]{
    \begin{tabular}{rllll}
    \toprule
    the number of training data &                  4 (0.03\%)  &                  7 (0.05\%)  &                  15 (0.1\%) &                  75 (0.5\%)\\
    \midrule
    Multilayer Perceptron      &  0.507 ($\pm$0.045) &  0.517 ($\pm$0.077) &    0.481 ($\pm$0.0) &  0.835 ($\pm$0.218) \\
    Random Forest       &  0.514 ($\pm$0.014) &   0.51 ($\pm$0.049) &  0.814 ($\pm$0.176) &  \textbf{0.974} ($\pm$0.005) \\
    Support Vector Classifier      &  0.557 ($\pm$0.068) &   0.61 ($\pm$0.191) &  0.737 ($\pm$0.177) &  0.963 ($\pm$0.013) \\
    \textbf{LassoMLP (ours)} &  \textbf{0.754} ($\pm$0.122) &  \textbf{0.771} ($\pm$0.118) &  \textbf{0.941} ($\pm$0.007) &  0.954 ($\pm$0.002) \\
    \bottomrule
    \end{tabular}
}%
\end{table*}

\begin{table*}[t]
\centering
\caption{Ratio (number and standard deviation) of remained features on MNIST784+Gauss5000  for running experiments 10 times.}
\label{tbl:Remained feature MNIST 2classes}
\subfigure[Classify 1 from 4\label{tbl:FS_MNIST1-4_table}]{
    \begin{tabular}{lllll}
    \toprule
    \#training data &                  4 (0.03\%)  &                  7 (0.05\%)  &                  15 (0.1\%) &                  75 (0.5\%)\\
    \midrule
    MNIST Features    &  0.021 (16.1$\pm$8.77) &  0.009 (7.0$\pm$1.63) &  0.004 (3.5$\pm$2.42) &  0.014 (10.8$\pm$16.7) \\
    Gaussian Features &   0.001 (4.3$\pm$3.33) &     0.0 (0.0$\pm$0.0) &     0.0 (0.0$\pm$0.0) &  0.001 (5.6$\pm$15.41) \\
    \bottomrule
    \end{tabular}
}%
\quad
\subfigure[Classify 1 from 7  
\label{tbl:FS_MNIST1-7_table}]{
    \begin{tabular}{lllll}
    \toprule
    \#training data &                  4 (0.03\%)  &                  7 (0.05\%)  &                  15 (0.1\%) &                  75 (0.5\%)\\
    \midrule
    MNIST Features    &  0.016 (12.2$\pm$9.09) &  0.007 (5.3$\pm$2.54) &  0.006 (5.0$\pm$5.44) &  0.006 (5.0$\pm$5.93) \\
    Gaussian Features &   0.001 (5.4$\pm$5.34) &     0.0 (0.0$\pm$0.0) &     0.0 (0.0$\pm$0.0) &     0.0 (0.0$\pm$0.0) \\
    \bottomrule
    \end{tabular}
}%
\end{table*}

\subsection{Generalization Performance by Feature Selection}
\label{sec:Experiments with MNIST}

\subsubsection{Setting}

% \begin{figure}[tb] % picture
%     \centering
%     \includegraphics[width=1.0\linewidth]{img/deep_lasso/llvsff.eps}
%     \caption{あいうえお}
%     \label{fig:mnist_precision}
% \end{figure}

% \input{fig_heatmap}

To show that feature selection can improve generalization performance, we compared the accuracy of LassoMLP with that of discriminators using all features in a setting with very little training data.
The task was binary classification, classifying 1 from 4 or 1 from 7 in MNIST.
%Some of the images used are shown in Fig.~\ref{fig:mnist13_images}.
We used as input a 5784-dimensional vector created by combining a 784-dimensional vector of digit images with a 5000-dimensional vector sampled from a standard Gaussian distribution.
The baseline methods are MLP, SVC, and RF.
The hyperparameters for LassoMLP are as follows: $\lambda=0.005$, $\rho=0.1$, $\delta=0.25$.
The batch size and the number of hidden units is 200 and 100 for both LassoMLP and MLP. The optimizer is SGD with the learning rate 0.1 for them.

\subsubsection{Result}

As shown in Table~\ref{tbl:accuracy_MNIST_2classes}, LassoMLP performed competitively with SVC and RF when the number of training data was 75, and outperformed SVC and RF when the number of training data was 15 or less, even though the discriminator was MLP. 

We believe that the reason why the accuracy of LassoMLP is relatively high even with a small number of training data is that the number of features is sufficiently smaller than the number of training data due to feature selection.
Indeed, as shown in Table~\ref{tbl:Remained feature MNIST 2classes}, the average number (ratio) of features used by LassoMLP ranges from 3.5 (0.4\%) to 16.1 (2.1\%). Furthermore, almost all irrelevant features of dimension 5000 have been removed.

\section{Conclusion}
\label{sec:conclusion}

We proposed a neural architecture  for embedded feature selection method, which is an important element technology of sparse modeling which attracts attention in recent years by extracting information from high dimensional data. The difference between the proposed method and the conventional neural network is that the first layer combines one-to-one and that the loss function includes the $L_1$ regularization term of the first layer weight, It is possible to naturally apply research results of deep learning which is actively studied.

We conducted several experiments to evaluate LassoMLP.
In the first experiment, we confirmed that LassoMLP works better for nonlinear dataset than vanilla Lasso.
In the second experiment, we showed that LassoMLP outperforms LassoNet in feature selection task, which means our method achieves state-of-the-art in our setting.
Lastly, we illustrated that the feature selection by LassoMLP improve the generalization performance through showing that LassoMLP predicts better than the full-feature classifier.
%the artificial data created from the additive model and the non-additive model was used to evaluate the feature selection appropriateness and the function approximation error, and the proposed method was better than Lasso.

In the future, we will discuss theoretical analysis of this method, comparison of performance with nonlinear methods, benchmark in real data such as gene microarray, application to neural networks with various structures such as convolutional neural network, stochastic gradient We plan to evaluate the performance in detail and expand the application range by using gradient method other than law etc.
% \input{introduction}
% \input{related_work}
% \input{deeplasso}
% \input{experiments}
% \input{exp_MNIST}
% \input{conclusion}

% Maybe, we should use bibtex...

\bibliographystyle{abbrv}
\bibliography{lasso_bib}

\end{document}